\documentclass{article}
\usepackage{amsmath}
\usepackage{amsfonts}
\usepackage{amssymb}
\usepackage{algorithm}
\usepackage{algpseudocode}
\usepackage{geometry}
\usepackage{graphicx}
\usepackage{tabularx}
\usepackage{booktabs} 
\usepackage{longtable}
\usepackage{hyperref}
\usepackage{authblk}
\usepackage{graphicx}
\usepackage{indentfirst}
\usepackage[utf8]{inputenc}
\DeclareUnicodeCharacter{2208}{$\in$}
\newcommand{\orcidd}[1]{\href{https://orcid.org/#1}{\includegraphics[width=10pt]{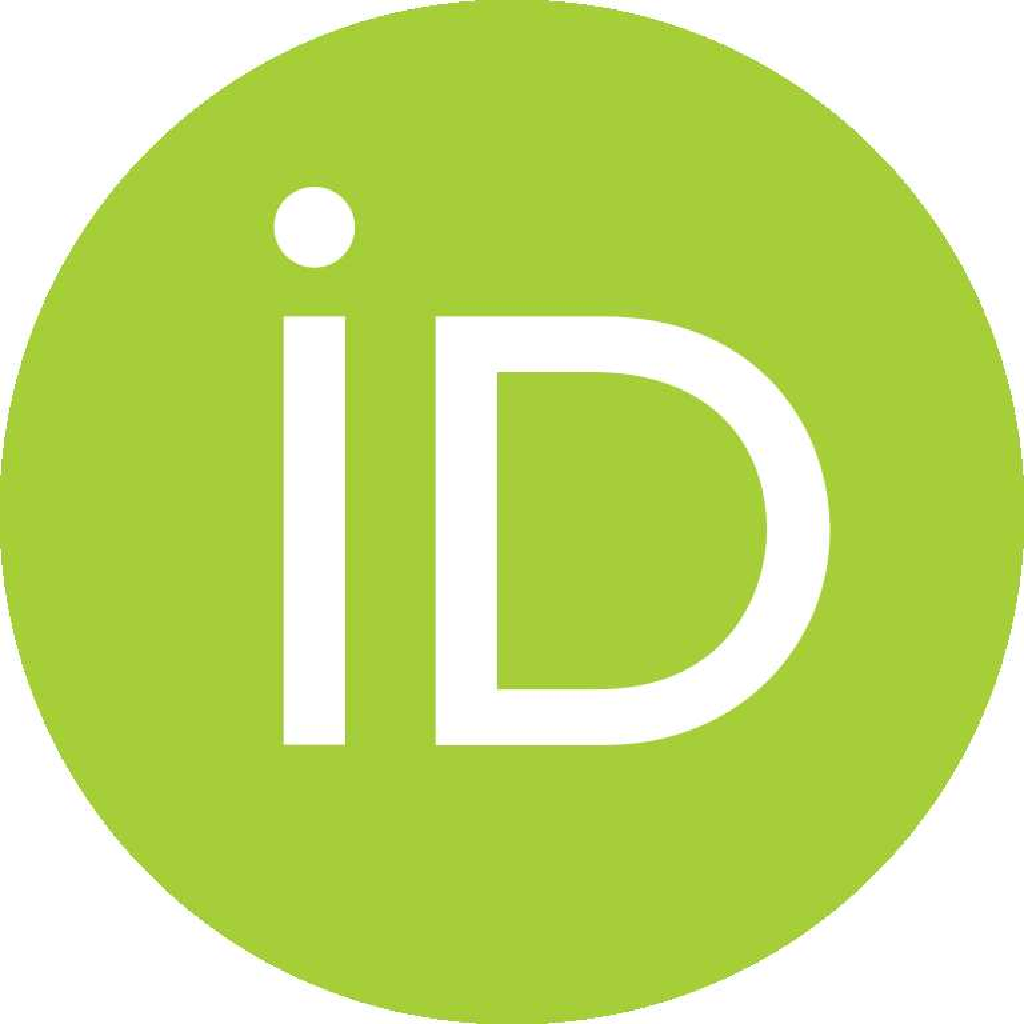}}}

\begin{document}

\title{A new simplified MOPSO based on Swarm Elitism and Swarm Memory: MO-ETPSO}
\author{Ricardo Fitas\thanks{Email: \href{mailto:ricardo.fitas@tu-darmstadt.de}{ricardo.fitas@tu-darmstadt.de}}\orcidd{0000-0001-5137-2451}}
\affil{Technical University of Darmstadt, Darmstadt, 64289, Germany}

\maketitle

\section*{Abstract}

This paper presents an algorithm based on Particle Swarm Optimization (PSO), adapted for multi-objective optimization problems: the Elitist PSO (MO-ETPSO). The proposed algorithm integrates core strategies from the well-established NSGA-II approach, such as the Crowding Distance Algorithm, while leveraging the advantages of Swarm Intelligence in terms of individual and social cognition. A novel aspect of the algorithm is the introduction of a swarm memory and swarm elitism, which may turn the adoption of NSGA-II strategies in PSO. These features enhance the algorithm's ability to retain and utilize high-quality solutions throughout optimization. Furthermore, all operators within the algorithm are intentionally designed for simplicity, ensuring ease of replication and implementation in various settings. Preliminary comparisons with the NSGA-II algorithm for the Green Vehicle Routing Problem, both in terms of solutions found and convergence, have yielded promising results in favor of MO-ETPSO.
\\

\textbf{Keywords:} Particle Swarm Optimization; NSGA-II; Green Vehicle Routing Problem; Swarm Memory; Swarm Elitism

\section{Introduction}

“Optimization” can be defined as the process of obtaining the maxima or minima in a set of available alternatives \cite{R015}.  Practically all engineering fields are impacted by optimization, which makes optimization wide-range \cite{R016}. Techniques for optimization are continuously evolving and improving \cite{R017}, so engineers are provided with increasingly sophisticated tools to enhance overall performance in various domains.

In the context of numerical optimization, an optimization problem can be defined mathematically. Let $N$ be the number of design variables and $\mathbf{X}$ = $\{x_1, \dots, x_N\}$ and $f(\mathbf{X})$ such an objective function. The number of inequality constraints is given by $n_g$ and the inequality constraints are $g_i(\mathbf{X}), \ i = 1, \dots, n_g$. Moreover, the number of equality constraints is given by $n_h$ and $h_j(\mathbf{X}), \ j = 1, \dots, n_h$, are the equality constraints. $\mathbf{X} \in \mathcal{S}^N$, where $\mathcal{S}^N = [x_{1_L}, x_{1_U}] \times \dots \times [x_{N_L}, x_{N_U}]$ is the search space. The standard form of a single-objective, constrained optimization problem is given in (\ref{eq1}).

\begin{equation}
\begin{aligned}
\text{Minimize: } & f(\mathbf{X}) \\
\text{Subject to: } & g_i(\mathbf{X}) \leq 0, \ i = 1, \dots, n_g \\
& h_j(\mathbf{X}) = 0, \ j = 1, \dots, n_h \\
& x_{k_L} \leq x_k \leq x_{k_U}, \ k = 1, \dots, N
\label{eq1}
\end{aligned}
\end{equation}

Objectives and constraints can be mathematically designed through explicit or implicit functions. Side constraints can be effectively handled through direct implementation. The optimal solution $\mathbf{X^*}$ involves identifying the combination that achieves the best objective function while satisfying all the equality, inequality, and side constraints.

When maximization problems occur, those can be transformed into minimization problems by inverting the signal of the objective function $f(\mathbf{X})$. Once looking at the optimal $\mathbf{X^*}$, one might be interested in $f(\mathbf{X^*})$, which represents the maximum value of $f$.

Engineering problems can also have multiple objectives. If this is the case, they are known as multi-objective optimization problems \cite{R200}. Those can be related to the minimization of more than one objective, maximization of them, or both situations in the same problem.

Two main methods for dealing with multi-objective optimization problems are Pareto dominance and scalarization \cite{R019}. Pareto dominance involves defining solutions that dominate all other solutions within the search space. This set of solutions is called the Pareto set. This also means that no element of the Pareto set can dominate any other element of it. Dominance is determined by a mathematical statement written in (\ref{eq2}), considering $\mathbf{X_1}, \mathbf{X_2} \in \mathcal{S}^N$~\cite{R020}.

\vspace{-6pt}
\begin{equation}
\mathbf{X_1} \prec \mathbf{X_2} \iff \forall i = 1, \dots, n_f : f_i(\mathbf{X_1}) \leq f_i(\mathbf{X_2}) \ \text{and} \ \exists i \in \{1, \dots, n_f\} : f_i(\mathbf{X_1}) < f_i(\mathbf{X_2})
\label{eq2}
\end{equation}

The Pareto set is a set with multiple valid solutions. One of the advantages of having a Pareto set is that the end user can choose an adequate solution. This can result in filtering a more extended list of combinations of possible solutions, where the filtered list is given to the final user. The set can also be used to calculate niches of solutions, which can have a tremendous impact on balancing exploitation and exploration. On the other hand, scalarization is an easier, sometimes more efficient approach, where various objectives are combined in a single fitness function to transform the problem into a single optimization problem. 

When it comes to solving optimization problems, there are various techniques available. The selection of a specific technique depends on its complexity and suitability for the problem at hand. Traditional methods, such as Newton's method and Quasi-Newton methods, involve calculating the evaluation function's gradient. However, modern engineering problems are often very complex. In this context, two major approaches are usually considered. 

A first approach is employing gradient-free algorithms, usually metaheuristics, which are more useful in such kinds of problems. These methods, including Genetic Algorithms (GA) \cite{R025}, Ant Colony Optimization (ACO) \cite{R026}, and Particle Swarm Optimization (PSO) \cite{R027}, among others, work by modifying instances of design variables, such as chromosomes in GA. In PSO, individual instances called particles move through the search space based on social and individual cognitions. Although both GA and PSO have convergence issues, variations, and alternatives have been published to address these problems. A second approach arises when surrogate models for the objective function become available or when metaheuristics prove to be computationally demanding.

In the field of multi-objective optimization,  Sharma and Kumar \cite{R028} examined various metaheuristic-based methods for optimizing multiple objectives. Perhaps the best-known Pareto dominance-based metaheuristic is the Non-dominated Sorting Genetic Algorithm II (NSGA-II) proposed by Deb et al. \cite{R050} in 2002. The novelty and authenticity of this algorithm rely on the criteria used to formulate the fitness function. While in scalarization, most of the fitness functions are a linear combination of the several objective functions, here there are two major criteria to define it: the rank attributed to a solution based on Pareto-based layers, and the crowding distance, where those solutions with higher distance to the other solutions are given more merit. This solution indirectly attributed more competition in the Pareto front region more efficiently, preventing stagnation and improving convergence.

Coello and Lechuga \cite{R051} also proposed a MOPSO algorithm in that same year. Despite the creation of an external "memory" in order to help with the adaptation of PSO to multi-objective, several authors have been suggesting other approaches for that adaptation - however, none of them were capable of considering a crowding distance or ranks. 

That was in 2005 when M. Reyes and C. Coello \cite{R052} proposed the OMOPSO algorithm, which showed promising results when compared to NSGA-II \cite{R055}. This algorithm was capable of replicating most of the NSGA-II features. However, NSGA-II was still and continued to be used in multiple works until now.

In 2022, Fitas et al. \cite{R053} designed a MOPSO algorithm with application in the design optimization problem of corrugated boards. The authors considered swarm memory and elitist strategies, but where NSGA-II features such as crowding distance and rank were considered. However, it was not compared directly with NSGA-II and not in benchmark problems such as the GVRP.

In this paper, a simplified MOPSO, based on the work from Fitas et al. \cite{R053}, is comprehensibly detailed and compared with NSGA-II using the known GVRP. The main advantage of the proposed algorithm, the Elitist PSO (MO-ETPSO), is that it captures the core features of NSGA-II but also considers the simplicity, ease of implementation, and continuous subspace handling capabilities of PSO compared to GA. 

The author hopes it can be applied to real-world problems. For now, this paper is focused on detailing the core strategy of the method and showing the core of its efficiency.

\section{Methodology}

\begin{figure}[ht]
  \centering
  \includegraphics[width=0.9\textwidth]{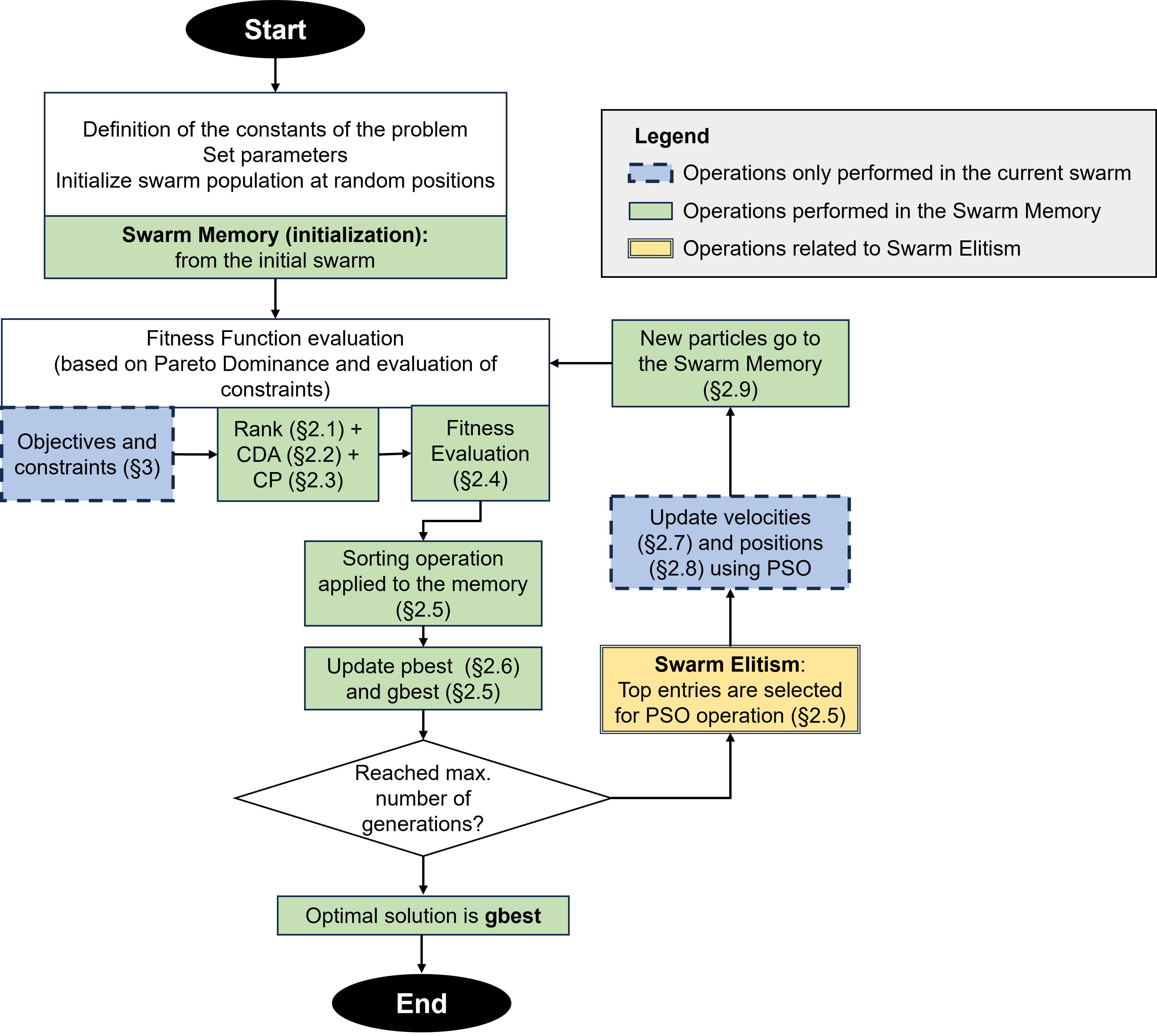}
  \caption{Flowchart for MO-ETPSO}
  \label{fig1}
\end{figure}

Figure \ref{fig1} presents the flowchart detailing the proposed MO-ETPSO's algorithmic process. Two novel concepts highlighted in the flowchart are 'Swarm Memory' (SM) and 'Swarm Elitism' (SE). SE is inspired by the elitist operator present in the NSGA-II algorithm, where the underlying merit is based on features such as crowding distance and rank related to the Pareto Dominance. 
SM results in a need for PSO to implement the elitism operator successfully. This need results from the fact that PSO operators have an individual and a social score, where a higher method exists for the swarm, but also for each particle. Serving as a repository for the swarm's collective experience, the SM operator is designed to retain information about the positions and qualities of the best solutions (particles) encountered. Its initialization phase is crucial as it sets the swarm's baseline to evolve. As the swarm's particles explore the solution space, their fitness is evaluated based on Pareto Dominance and the assessment of constraints.

On the other hand, SE is a mechanism that ensures the highest-quality solutions are retained over successive generations. It operates by selecting the top entries from the swarm memory for use in PSO operations, thereby biasing the swarm towards regions of the solution space that have previously yielded the best results among the swarm. This approach is similar to evolutionary strategies that favor the survival of the fittest, thus continuously refining the swarm's search towards the optimal solution. The algorithm updates the velocities and positions of the particles using PSO rules for the selected SE, with each new generation of particles potentially contributing to the swarm memory.

The process iterates in the following way: the particles in the swarm are evaluated using the function objectives. Then, all the history and the current swarm are compared, where rank, crowind distance algorithm, and constraints are evaluated. Then, a final expression is assessed, the "Fitness Evaluation". After sorting the entries, the 'pbest' and 'gbest' are updated and the top entries are selected and operated using PSO equations for both velocity and position, so that the new particles are once again evaluated. This happens until a stopping criterion is met. In the current work, that stopping criterion is the maximum number of generations. At this point, the optimal solution is identified as the 'gbest,' representing the global best solution found by the swarm, being 'pbest' the representation of the individually best solution for each particle.

\subsection{Rank Attribution Process}

The rank attribution process begins by considering a population \( P \) of particles. Each particle \( p_i \) is associated with a vectorial objective function \( F(p_i) \).

\begin{figure}[ht]
  \centering
  \includegraphics[width=0.8\textwidth]{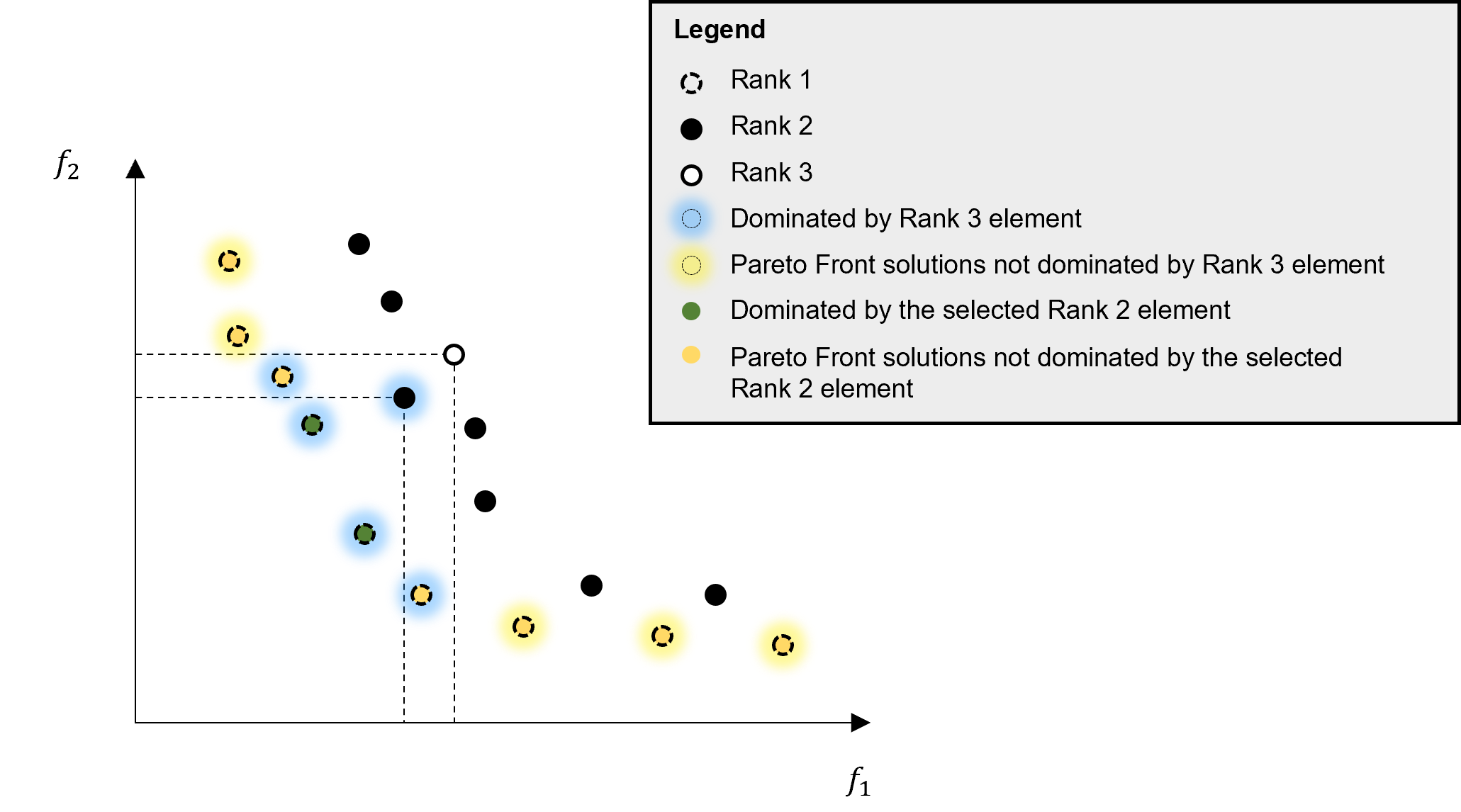}
  \caption{Illustration of the Rank Attribution}
  \label{fig2}
\end{figure}

The non-dominated sorting approach is employed to rank particles based on Pareto dominance. Particles are then categorized into different Pareto fronts. The first front, \( F_1 \), comprises particles that are not dominated by any other member of the population. These particles are assigned the highest priority, indicating their proximity to the Pareto optimal front. Figure \ref{fig2} illustrates the different rank attributions.

To determine subsequent fronts, the algorithm iteratively removes the identified front from the population and reapplies the non-dominated sorting to the remaining particles. This process iterates, with the rank increasing sequentially, until all particles have been classified into fronts.

The mathematical representation of the iterative non-dominated sorting is as follows:

\begin{algorithm}
\caption{Non-dominated Sort Algorithm}
\begin{algorithmic}[1]
\State Let \( P \) be the initial population of particles.
\State Initialize front counter \( n \) to 1.
\While{there are particles left in \( P \)}
    \For{each particle \( p_i \in P \)}
        \If{\( p_i \) is non-dominated}
            \State Assign \( p_i \) to front \( F_n \).
        \EndIf
    \EndFor
    \State Remove particles in \( F_n \) from \( P \).
    \State Increment \( n \).
\EndWhile
\end{algorithmic}
\end{algorithm}

The ranks ascribed to the particles serve a dual purpose. Firstly, they act as a mechanism to discriminate between the quality of solutions, favoring those closer to the Pareto optimal front. Secondly, they guide the swarm's search process, steering it towards regions of the search space with higher potential for optimality.

\subsection{Crowding Distance Computation}
After the particles are sorted into different Pareto fronts using the rank attribution mechanism, crowding distance computation is used to maintain diversity in the population. The crowding distance for a particle is a measure of how close it is to its neighbors. A larger crowding distance implies a less crowded neighborhood, which is preferred to ensure a diverse set of solutions. Here, the concept of the crowding distance considers the volume of the formed hypercube due to the solutions in the neighborhood, rather than a relative distance.

For each front \( F_k \) obtained from the rank attribution process, the crowding distance for each particle in the front is calculated. The crowding distance of a particle \( p_i \), \( D(p_i) \), is the product of the side lengths of the hypercube formed by its adjacent particles in the objective space. Mathematically, it is defined in (\ref{eq3}).

\begin{equation}
D(p_i) = \prod_{m=1}^{M} \left( \frac{f_m(p_{i+1}) - f_m(p_{i-1})}{\max(F_k^{(m)}) - \min(F_k^{(m)})} \right)
\label{eq3}
\end{equation}

\begin{figure}[ht]
  \centering
  \includegraphics[width=0.8\textwidth]{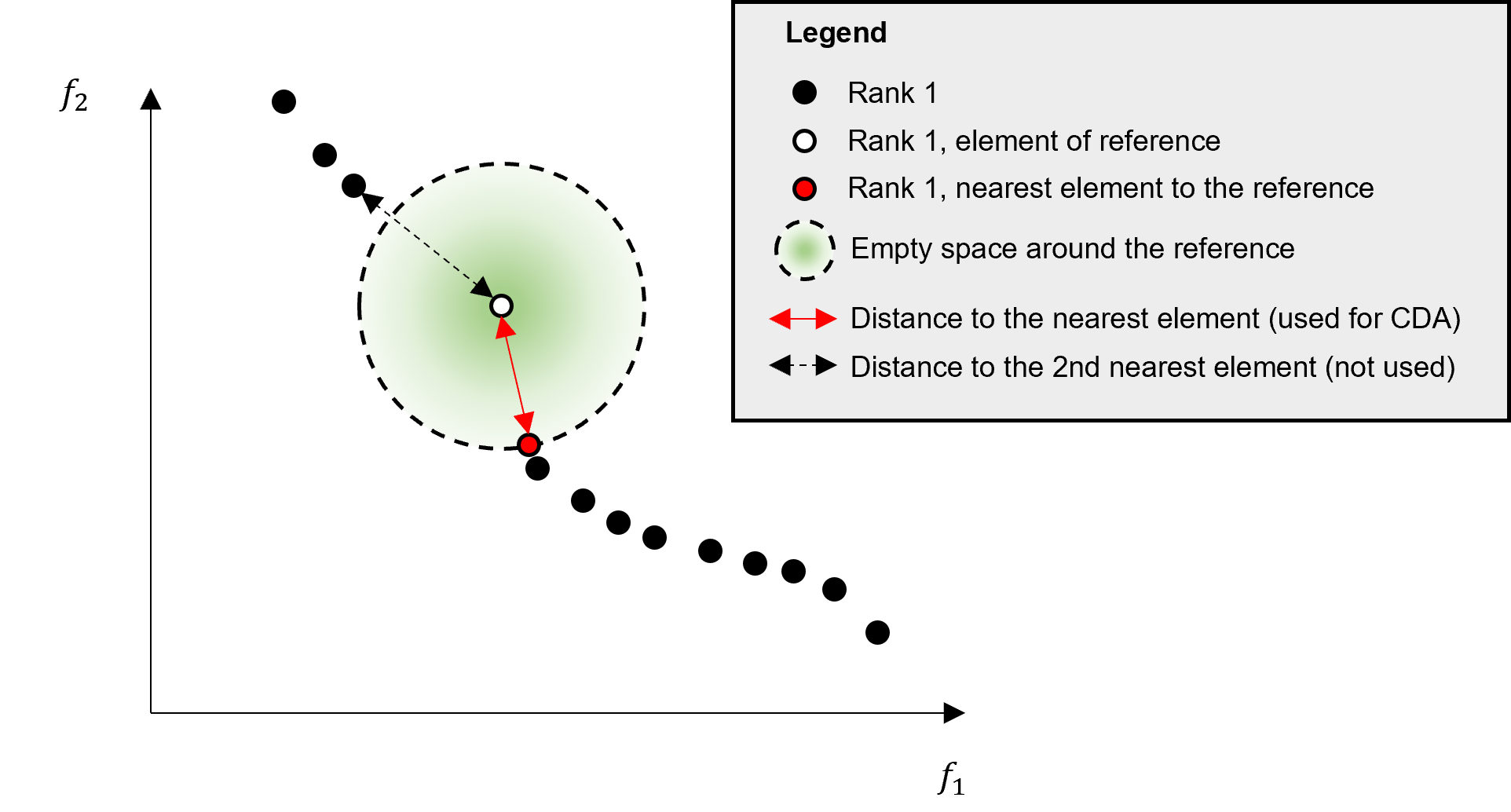}
  \caption{Illustration of the Crowding Distance}
  \label{fig3}
\end{figure}

Here, \( f_m(p) \) denotes the \( m^{th} \) objective value of particle \( p \), and \( \max(F_k^{(m)}) \) and \( \min(F_k^{(m)}) \) are the maximum and minimum values of the \( m^{th} \) objective in front \( F_k \), respectively. Particles at the boundaries of the front are assigned an infinite crowding distance, ensuring their selection for the next generation. Figure \ref{fig3} illustrates the crowding distance associated with a solution of the objective space.

The computation of crowding distance is integrated into the algorithm as follows:

\begin{algorithm}
\caption{Crowding Distance Computation}
\begin{algorithmic}[1]
\For{each front \( F_k \)}
    \For{each objective \( m \)}
        \State Sort \( F_k \) based on \( f_m \) values.
        \State Assign infinite crowding distance to boundary particles in \( F_k \).
        \For{each particle \( p_i \) in \( F_k \) except boundary particles}
            \State Compute crowding distance \( d_i \) as per (\ref{eq3}).
        \EndFor
    \EndFor
\EndFor
\end{algorithmic}
\end{algorithm}

The crowding distance computation ensures that the selection process is not only particles that are close to the Pareto front but also those that are isolated in the objective space, thereby maintaining diversity in the swarm.

\subsection{Constraints Penalty}
In multi-objective optimization, constraints define the feasibility of solutions. The penalty for constraint violation is incorporated into the fitness evaluation, discouraging the selection of infeasible solutions. For a particle \( p_i \), let \( C(p_i) \) denote the number of constraints violated. The penalty reduces the fitness of solutions violating constraints.

\subsection{Final Merit}
The final merit \( F(p_i) \) of a particle \( p_i \) combines its rank, crowding distance, and penalty for constraints violation, formulated as in (\ref{eq4}).

\begin{equation}
F(p_i) = \max_j R(p_j) - R(p_i) + D(p_i) - C(p_i)
\label{eq4}
\end{equation}

Here, \( R(p_i) \) is the rank, \( D(p_i) \) is the crowding distance, and \( C(p_i) \) is the constraint violation penalty. The idea is to maximize the final merit, by minimizing the rank of the particle, maximizing the crowding distance, and minimizing the number of violated constraints.

\subsubsection{Merit Hierarchy Proof}
The fitness function is designed to prioritize non-violating, Pareto-efficient solutions with good crowding distances. We prove this as follows:

\begin{enumerate}
    \item Non-violating solutions are prioritized since \( C(p_i) \) directly reduces fitness for violating solutions.
    \item Lower ranks (closer to Pareto front) yield higher fitness due to the direct addition of \( R(p_i) \) to fitness.
    \item Higher crowding distances are favored within the same rank due to the term \( D(p_i) \).
    \item Violating solutions cannot surpass non-violating solutions in fitness, regardless of rank or crowding distance.
    \item Solutions not on the Pareto front do not significantly benefit from the crowding distance term, as their rank places them higher in merit. It is possible to verify that, due to the nature of the expressions, $D(p_i) \leq 1 \leq \min_j R(p_j)$.
\end{enumerate}

\subsection{Swarm Memory and Elitism}

SM is initialized as an empty collection at the start, ready to be populated with the swarm's data as the algorithm progresses. Throughout the optimization process, each particle's information is recorded in the memory, which includes a set of attributes reflecting the state and quality of each solution candidate. The new entries also contain the information from the particle that found those entries. This is crucial information because it allows not only the tracking of the particles but also the calculation of the best solution for each particle.

The list of labels for the swarm memory includes an identifier for each particle, the particle's position in the solution space, its velocity, and the objectives or fitness values representing how well the particle's current position satisfies the optimization criteria. Figure \ref{fig4} below illustrates the swarm memory.

\begin{figure}[ht]
  \centering
  \includegraphics[width=\textwidth]{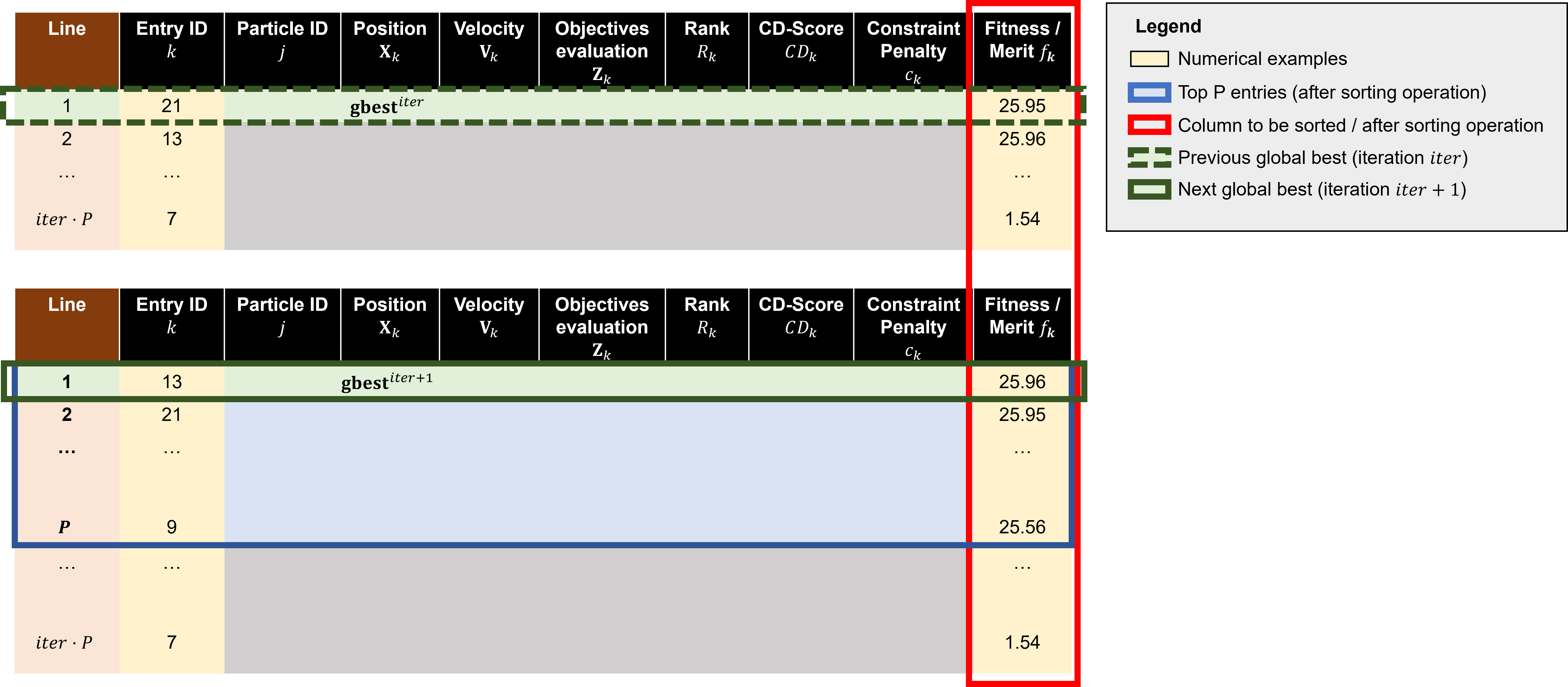}
  \caption{Swarm Memory and Elitism}
  \label{fig4}
\end{figure}

Swarm elitism is manifested in the sorting operation by merit, which rearranges the particles in descending order of their merit scores. This operation ensures that the most promising particles, those with the highest merit—typically reflecting a lower rank (closer to the Pareto front), a higher crowding distance (indicating less crowded neighborhoods), and no constraint violations—are placed at the top of the list. The particle with the highest merit score at the end of an iteration is considered the 'gbest' (global best) for the next iteration.

The table in the bottom captures the swarm after the sorting operation. After the first iteration, for instance, the Entry IDs are no longer in ascending numerical order, because the particles have been repositioned based on merit rather than their original order. The top $P$ entries are selected for further application of the PSO equations of velocity of position. However, before going to this, the 'pbest' is still to be updated for each particle.

\subsection{Choosing the personal best}

The personal best is a characteristic of PSO that does not exist in NSGA-II or related algorithms. Nevertheless, it also means that adopting elitism in PSO might be more challenging. In this paper, the 'pbest' for each particle is then updated based on the following strategy.

\begin{figure}[ht]
  \centering
  \includegraphics[width=0.7\textwidth]{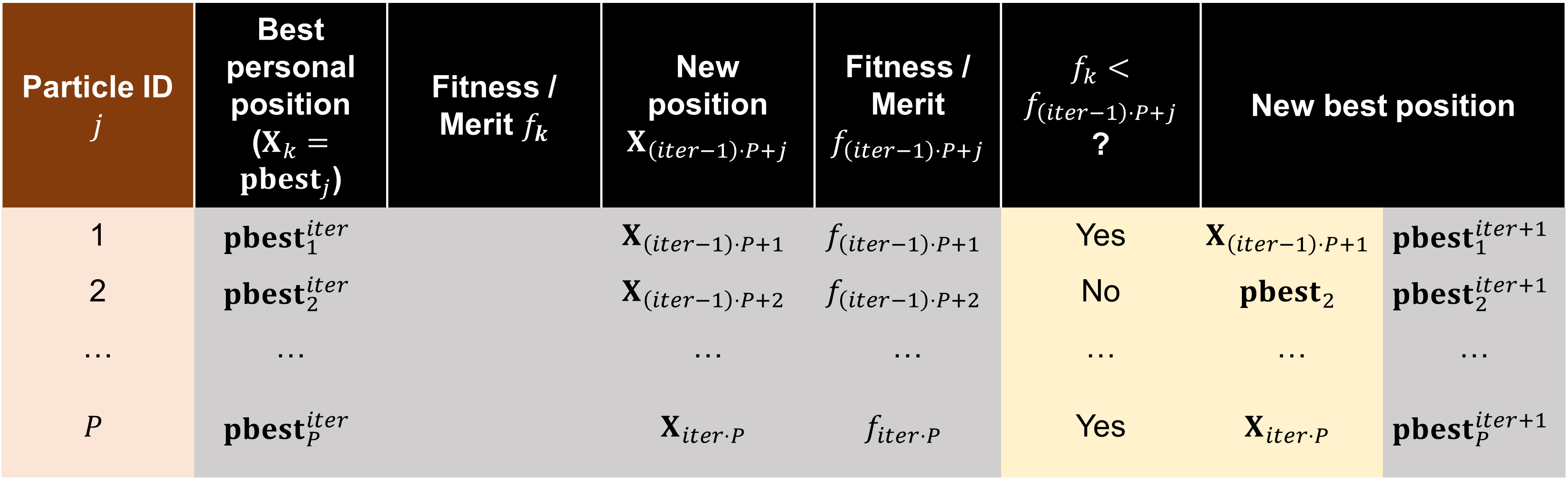}
  \caption{Process of choosing personal best}
  \label{fig5}
\end{figure}

The table in Figure \ref{fig5} shows, for each particle identified by Particle ID, its best personal position ('pbest') found until the current iteration (iter), the new position it has explored in the current iteration, the fitness/merit of this new position, and a decision on whether the new position becomes the particle's new personal best for the next iteration.

The column Fitness / Merit represents the computed merit of the new position. The decision on whether the new position becomes the new personal best is based on whether the merit of the new position is greater than that of the current personal best, tracked by the particle ID. If the merit is higher (Yes), the new position is accepted as the new personal best; if not (No), the particle retains its previous personal best.

This table maintains a constant number of entries across iterations, in contrast to the swarm memory, which might vary in size, depending on the memory limitations for speed improvement. Each entry corresponds to a unique particle, ensuring the personal best is tracked consistently over time. The swarm memory is leveraged to filter out solutions by particle ID, identifying the solution with the highest merit as the personal best for that particle.

\subsection{Velocity Update}
In PSO, the velocity update rule is the mechanism that guides the swarm toward optimal regions in the search space. For each particle \( i \), the velocity update is given in (\ref{eq5}.

\begin{equation}
\mathbf{v}_{i}^{new} = K \left[ w \mathbf{v}_{i} + c_1 r_1 (\mathbf{pbest}_{i} - \mathbf{x}_{i}) + c_2 r_2 (\mathbf{gbest} - \mathbf{x}_{i}) \right]
\label{eq5}
\end{equation}
where:
\begin{itemize}
\item \( \mathbf{v}_{i} \) is the current velocity of the particle.
\item \( \mathbf{pbest}_{i} \) is the personal best position of the particle.
\item \( \mathbf{gbest} \) is the global best position among all particles.
\item \( \mathbf{x}_{i} \) is the current position of the particle.
\item \( w \) is the inertia weight that controls the impact of the previous velocity of the particle on its current one.
\item \( c_1 \) and \( c_2 \) are the cognitive and social coefficients, respectively, which represent the private and collective influence on the velocity update.
\item \( r_1 \) and \( r_2 \) are random numbers uniformly distributed in [0, 1].
\item \( K \) is the constriction factor that ensures the convergence of the swarm, calculated as in (\ref{eq6}).
\begin{equation}
K = \frac{2}{|2 - \phi - \sqrt{\phi^2 - 4\phi}|}
\label{eq6}
\end{equation}
with \( \phi = c_1 + c_2 \), ensuring that \( \phi > 4 \) to provide a converging swarm behavior.
\end{itemize}

\subsection{Position Update}
Following the velocity update, the position of each particle is adjusted accordingly. The updated position must adhere to the problem's side constraints, which typically include the search space bounds. The position update is performed as in (\ref{eq7}).
\begin{equation}
\mathbf{x}_{i}^{new} = \max(\mathbf{bound}_{i}^{min}, \min(\mathbf{bound}_{i}^{max}, \mathbf{x}_{i} + \mathbf{v}_{i}^{new}))
\label{eq7}
\end{equation}
For continuous search spaces, this ensures that the particle remains within the defined bounds. In the case of discrete search spaces, the position is discretized appropriately, often by rounding to the nearest valid discrete value.

A numerical example of velocity and position update in PSO is presented in \cite{R200}.

\subsection{Assembling New Entries}
\begin{figure}[ht]
  \centering
  \includegraphics[width=\textwidth]{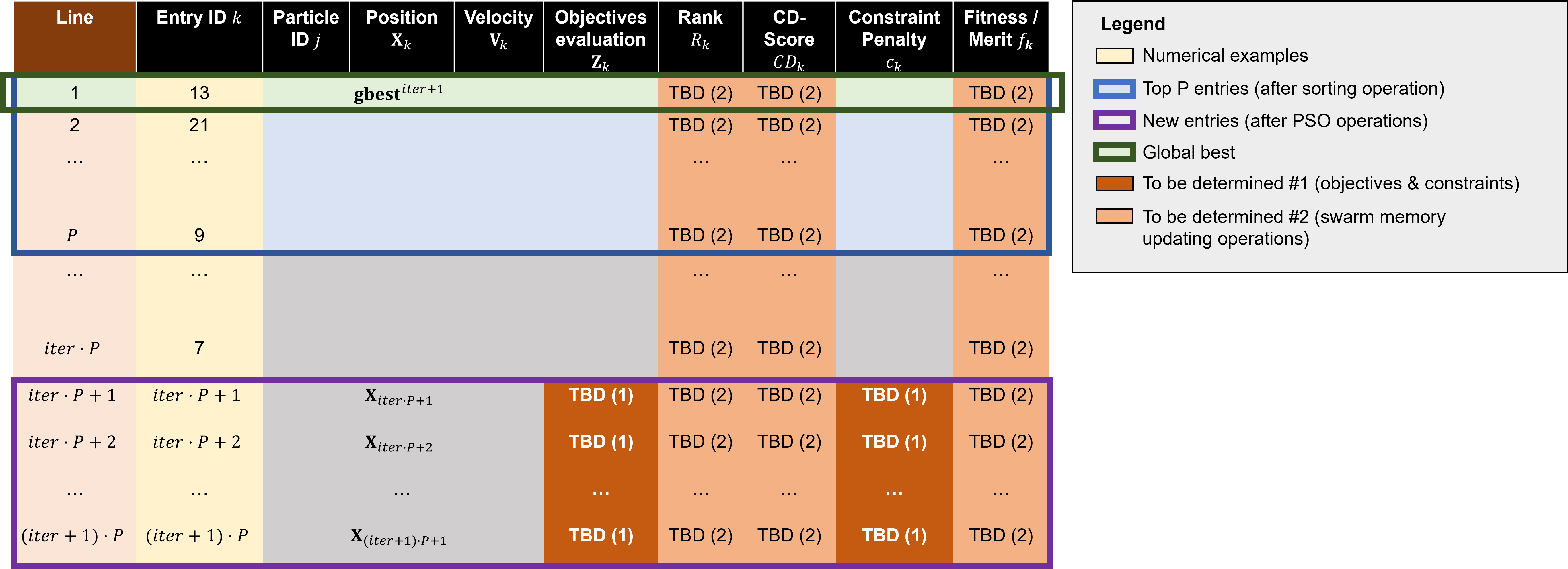}
  \caption{Assembling new entries in Swarm Memory}
  \label{fig6}
\end{figure}

Upon completion of a PSO iteration, the algorithm incorporates new candidate solutions into the swarm memory. These new entries, derived from the top-performing particles, are inserted while preserving the indexing order, as Figure \ref{fig6} shows. This ensures a structured approach to subsequent operations, allowing for a systematic evaluation of the new solutions.

After this step, two types of information are now to be determined: the objectives and constraints, from the evaluation of the objectives, and the merit-related components. For that, all the process aforementioned in the subsections are to be repeated until the convergence is met.

\subsection{Convergence Improvement}

In metaheuristics, convergence-related issues due to the stagnation of the process and due to the incorrect balance between exploration and exploitation, are not uncommon. MO-ETPSO distinguishes itself from other metaheuristics due to its capability of global convergence, mathematically shown by \cite{R040}. The following topics discuss how MO-ETPSO is capable of improving convergence.

The algorithm maintains a stagnation counter, incremented each time the Pareto front remains unchanged post an iteration. When this counter surpasses a predefined threshold, the algorithm initiates a randomization phase for a subset of the swarm. Here, the selection of particles is random, with the number of particles being a function of the stagnation counter.

The randomization process is formally defined as follows: For a swarm of particles \( S \) and a stagnation counter \( \gamma \), a subset \( S_{rand} \subset S \) is randomly selected. The positions of particles in \( S_{rand} \) are reinitialized within the search space bounds. This process is mathematically represented as in (\ref{eq8}).
\begin{equation}
\mathbf{x}_{i}^{new} = \text{rand}(\mathbf{bound}_{i}^{min}, \mathbf{bound}_{i}^{max}), \quad \forall i \in S_{rand}
\label{eq8}
\end{equation}
where \( \text{rand}(a, b) \) denotes a uniform random sampling within the interval \([a, b]\).

Following the randomization, the stagnation counter is reset to zero. The PSO then resumes its iterative process of optimization. The randomization step ensures that the swarm does not become trapped in local Pareto fronts and continues to explore the search space effectively.

\section{Numerical Example: Green Vehicle Routing Problem (GVRP)}
MO-ETPSO will be tested on the GVRP. The goal will be the optimization of routes considering factors like speed, load, distance, and emissions. 

\begin{figure}[ht]
  \centering
  \includegraphics[width=\textwidth]{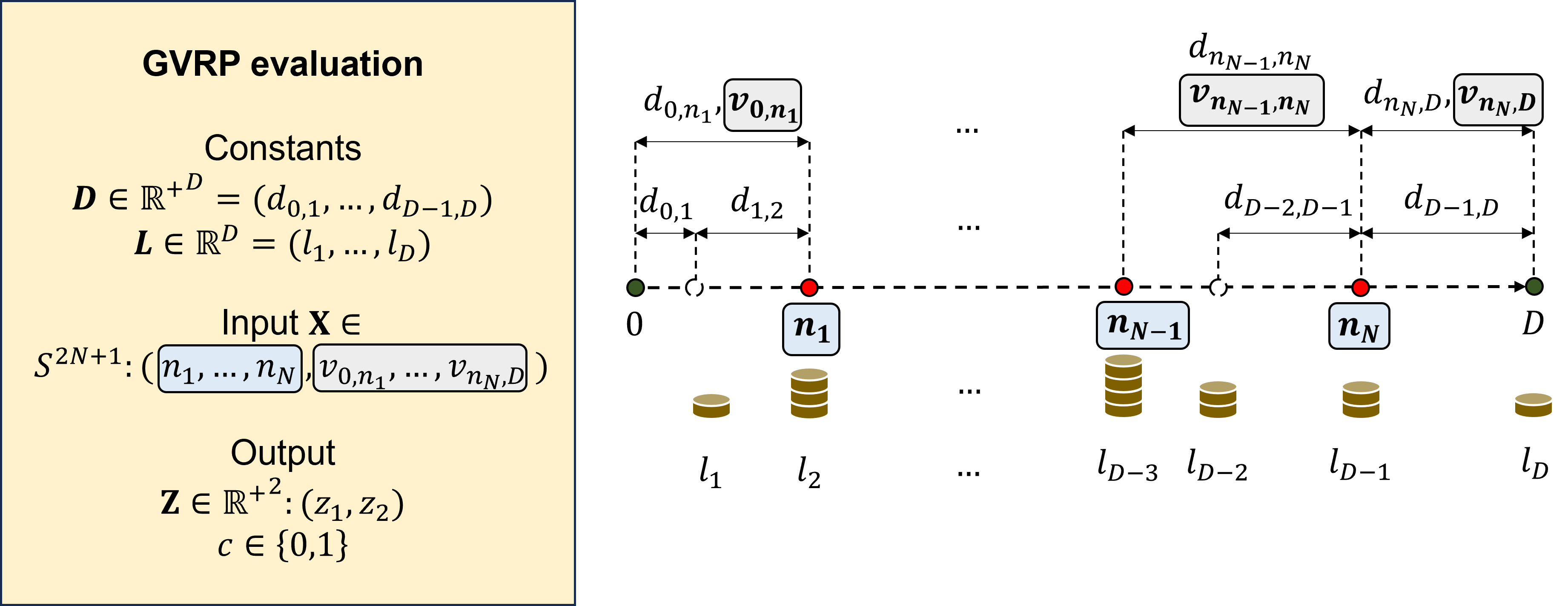}
  \caption{Gree Vehicle Routing Problem}
  \label{fig7}
\end{figure}

Figure \ref{fig7} illustrates the problem. In the given example, the vehicle starts from the node with index $n_0 = 0$, going directly, in this case, to the node with index $n_1 = 2$, where the vehicle has a load $L_1 = L_0 + l_2$, and not stopping in node number 1. At the end, the vehicle decides to stop in the node with index $D-1$, going directly to the last node, with index $D$. The green dots represent the starting and ending nodes, and the red ones represent the nodes where the vehicle stops.

The constants of the problem are the distances $d$ between the nodes $i$ and $i+1$, with $i = 0, ..., D$, and the expected increments $l_i$ related to the vehicle's capacity. These increments can also be negative, meaning the vehicle is shedding weight. The inputs are the indices of the nodes $n_i$, with $i = 0 ... N+1$ where the vehicle stops (starting from the index $n_0 = 0$ and ending in the index $n_{N+1} = D$), followed by the velocities between the nodes $n_{i-1}$ and $n_i$. The outputs used for the optimization algorithm's feed are the two objectives $f_1 = z_1$ and $f_2 = z_2$ and the constraint $c$.

The vehicle capacity is updated in the node $n+i$, using (\ref{eq9}).

\begin{equation}
L_{n_{i+1}} = L_{n_{i}} + l_{n_{i+1}}
\label{eq9}
\end{equation}

The two objectives are defined as follows:
\begin{enumerate}
    \item Minimize the overall travel time:
    \begin{align}
    z_1 &= \sum_{i=0}^{N} \left( \frac{d_{n_i,n_{i+1}}}{v_{n_i,n_{i+1}}} \right)
    \label{eq10}
    \end{align}
    \item Minimize the emissions considering the load and speed of the vehicle:
    \begin{align}
        z_2 &= \sum_{i=0}^{N} \left(d_{n_i,n_{i+1}} \cdot L_{n_i} \cdot (0.05\cdot v_{n_i,n_{i+1}} + 1)  \right) \times 10^{-10}
        \label{eq11}
    \end{align}
\end{enumerate}

The constraint $c$ is defined in (\ref{eq12}).

\begin{equation}
c = \begin{cases} 
1 & \text{if } \exists i: L_{n_i} > Q \vee L_{n_i} < 0, i = 0 \dots {N+1}\\
0 & \text{otherwise}.
\end{cases}\label{eq12}
\end{equation}

Besides the side constraints, the indices are sampled so that the difference between consecutive nodes is neither below zero (meaning the vehicle does not go back) nor above D/N. This prevents too many potential solutions from being discarded in the iterative process. The real vector in the search space is structured in (\ref{eq13}).
\begin{equation}
\mathbf{X} = (x_1 = \lceil\bar{x}_1 k_S\rceil, ..., x_N = \lceil\bar{x}_N k_S\rceil, v_{0,x_1}, ..., v_{x_N,D})\\
\label{eq13}
\end{equation}
\begin{equation}
k_S = \frac{D-1}{\bar{x}_N}\\
\label{eq14}
\end{equation}
Where $\bar{x}_{i+1} - \bar{x}_i \in ]0,D/N[$. The multi-objective optimization problem is defined in (\ref{eq15}).

\begin{equation}
\begin{aligned}
\text{Minimize: } & F(\mathbf{X}) = (z_1, z_2) \\
\text{Subject to: } & c(\mathbf{X}) = 0 \\
& 0 < \bar{x}_{k+1} - \bar{x}_k < D/N \wedge x_k \in \mathbb{N} , \ k = 1, \dots, N \\
& v_{min} \leq x_k \leq v_{min}, \ k = N+1, \dots, 2N+1
\end{aligned}\label{eq15}
\end{equation}

\section{Results and Discussion}

\subsection{NSGA-II Setup}

In the implementation of NSGA-II, the DEAP (Distributed Evolutionary Algorithms in Python) library is utilized. DEAP is a versatile evolutionary computation framework, which offers extensive support for genetic algorithms, including NSGA-II. It provides a range of tools and functions that facilitate the customization of the genetic algorithm's workflow, including selection, crossover, and mutation processes, as well as the handling of multiple objectives and constraints.

The selection process in NSGA-II, as implemented through DEAP's tools.selNSGA2 function. Crossover and mutation are genetic operators in NSGA-II, used to generate new candidate solutions. In this implementation, a two-point crossover (tools.cxTwoPoint) is employed, where segments of parent chromosomes are exchanged to produce offspring. Mutation carried out through Gaussian mutation (tools.mutGaussian), introduces random changes to the individuals, helping to explore new areas of the solution space and preventing the algorithm from getting stuck in local optima. The parameters for mutation are carefully chosen to balance the exploration and exploitation aspects of the search.

Alongside NSGA-II, a Hall of Fame (HoF) is used to store the best solutions found during the evolutionary process. In the context of multi-objective optimization, the Hall of Fame typically represents a Pareto front, a collection of non-dominated solutions that showcase the trade-offs between the different objectives.

\subsection{Experimental Setup}

In this paper, MO-ETPSO is compared with NSGA-II. Therefore, Both NSGA-II and MO-ETPSO are run for the GVRP until convergence. The parameters are defined for the problem, as follows: $Q = 200$, $v_{min} = 1$, $v_{max} = 100$. Moreover, there are two different test sets for this problem: the total number of nodes the vehicle can stop, $N$, is equal to 10 in the first one, and equal to 100 in the second problem. This means that there are also 10 and 100 different velocities, respectively. While multiple problems can handle relatively small-to-medium search spaces ($2N = 20$), the second one is built in order to show that MO-ETPSO was also designed to handle highly-dimensional search spaces ($2N = 200$). Also, the first problem has $D = 1000 + 1$ nodes, whereas the second one has $D = 1000000 + 1$. The mapping and the extra loading are given in the following repository: https://github.com/ricardofitas/MO-ETPSO-GVRP.

Preliminary executions of the algorithm allowed for the verification that convergence for MO-ETPSO happens around 100 iterations, while the convergence for NSGA-II happens at the end of 50 iterations. Therefore, both MO-ETPSO and NSGA-II are run for 100 iterations each run. Both NSGA-II and MO-ETPSO were run 16 times each.

For the MO-ETPSO parameters, the number of particles was set to $n_{\text{particles}} = 100$. The inertia weight $w$ was chosen as $0.7$, and the coefficients $c_1$ and $c2$ were both set to $2.05$.

The parameters chosen for NSGA-II were the following: the mutation operator \texttt{tools.mutGaussian} was configured with a mean $\mu = 0$, standard deviation $\sigma = 1$, and the probability of mutating an individual gene \textit{indpb} = 0.1. The population size was set to $\text{MU} = 20$, and the offspring size, denoted as $\Lambda$, was $80$. The crossover probability (\textit{CXPB}) and mutation probability (\textit{MUTPB}) were set to $0.7$ and $0.2$, respectively. For each iteration, this configuration necessitates a certain number of evaluations, specifically $\text{MU} + \Lambda = 100$ evaluations to account for both the current population and the generated offspring.

\subsection{Results}

Table \ref{tab1} presents the list of points (two first columns) belonging to the Pareto front, for MO-ETPSO and Problem \#1. The corresponding entry from the Search Space is also given. It is possible to observe that the time is minimized (first rows) by maximizing the velocity throughout the path while minimizing the velocity will lead to higher times but fewer emissions (last rows).

\begin{longtable}{>{\raggedleft\arraybackslash}p{2cm} >{\raggedleft\arraybackslash}p{1cm} p{13cm}}
\caption{MO-ETPSO Pareto front solutions} \label{tab1} \\
\hline
$\mathbf{Z_i^{(1)}}$ & $\mathbf{Z_i^{(2)}}$ & \multicolumn{1}{c}{$\mathbf{X_i}$} \\ \hline
\endfirsthead
\multicolumn{3}{c}%
{{\bfseries \tablename\ \thetable{} -- continued from previous page}} \\
\hline $\mathbf{Z_i^{(1)}}$ & $\mathbf{Z_i^{(2)}}$ & \multicolumn{1}{c}{$\mathbf{X_i}$} \\ \hline
\endhead
\hline \multicolumn{3}{r}{{Continued on next page}} \\ \hline
\endfoot
\hline
\endlastfoot
5199.5298 & 0.5398 & 3, 110, 112, 336, 495, 719, 942, 945, 1000, 100, 100, 100, 100, 100, 100, 100, 100, 100, 1 \\ 
5199.5298 & 0.3810 & 2, 4, 180, 182, 357, 532, 686, 861, 999, 100, 100, 100, 100, 100, 100, 100, 100, 100, 100 \\ 
5204.5666 & 0.3241 & 4, 148, 152, 340, 344, 646, 727, 892, 999, 100, 100, 100, 100, 66.2731, 100, 100, 100, 100, 100 \\ 
5211.6731 & 0.2503 & 3, 144, 147, 339, 342, 612, 762, 963, 999, 100, 100, 100, 99.9992, 45.3406, 100, 100, 100, 100, 100 \\ 
5229.3240 & 0.2343 & 3, 146, 149, 339, 342, 610, 774, 997, 999, 100, 100, 100, 98.6058, 39.8787, 100, 100, 100, 100, 100 \\ 
5237.4329 & 0.2169 & 3, 167, 169, 335, 338, 591, 746, 997, 999, 100, 100, 100, 100, 29.2028, 100, 100, 100, 100, 100 \\ 
5260.3957 & 0.1885 & 4, 145, 148, 336, 339, 610, 752, 957, 999, 100, 100, 100, 95.8858, 47.4258, 100, 100, 100, 100, 100 \\ 
5591.7718 & 0.1815 & 76, 80, 84, 88, 448, 452, 636, 640, 999, 100, 92.1036, 100, 100, 100, 100, 100, 67.5343, 100, 1 \\ 
6252.0563 & 0.1800 & 51, 144, 148, 361, 365, 622, 749, 816, 1000, 91.4405, 96.3137, 100, 74.1968, 69.3901, 100, 89.2088, 100, 64.0105, 83.5669 \\ 
6276.7788 & 0.1767 & 5, 7, 180, 355, 367, 524, 663, 836, 999, 43.4773, 23.3517, 67.5026, 95.2021, 95.8681, 88.9251, 80.7612, 79.6662, 96.6061, 100 \\ 
6532.1522 & 0.1690 & 5, 7, 175, 341, 370, 536, 683, 849, 999, 49.9078, 36.7793, 65.3946, 82.598, 94.9492, 82.1308, 80.6428, 80.849, 93.8114, 100 \\ 
6727.4640 & 0.1629 & 24, 114, 153, 381, 495, 704, 886, 901, 999, 51.402, 73.3662, 71.3998, 73.8406, 82.3198, 76.8115, 81.4062, 93.1413, 91.9379, 5.7764 \\ 
7280.3151 & 0.1381 & 77, 133, 144, 266, 545, 587, 706, 748, 999, 81.2198, 33.4432, 94.1152, 68.4098, 80.1851, 60.3523, 90.3703, 56.8639, 79.2012, 56.6845 \\ 
8544.8688 & 0.1380 & 18, 219, 266, 366, 587, 590, 653, 874, 999, 50.0608, 46.2992, 39.1442, 54.9674, 89.7272, 82.7368, 97.8985, 58.1299, 76.189, 64.2512 \\ 
8943.7451 & 0.1169 & 64, 145, 148, 350, 353, 572, 732, 855, 999, 42.7833, 50.4361, 100, 53.6965, 27.4764, 100, 52.8713, 68.7273, 45.9207, 55.7181 \\ 
12543.9038 & 0.0967 & 3, 6, 170, 250, 439, 689, 877, 997, 999, 33.5209, 100, 37.4876, 23.4178, 73.5869, 34.5804, 50.1895, 48.4265, 99.0647, 100 \\ 
16550.2975 & 0.0898 & 23, 116, 265, 319, 427, 618, 689, 833, 1000, 46.9961, 31.5618, 46.4423, 62.6331, 26.5533, 28.527, 48.2595, 55.574, 17.0004, 78.8157 \\ 
16916.4953 & 0.0870 & 3, 6, 154, 290, 402, 654, 846, 997, 999, 16.8336, 100, 47.1258, 55.8964, 79.2401, 28.6552, 15.6433, 38.7029, 78.1274, 100 \\ 
17037.3548 & 0.0869 & 3, 6, 154, 290, 402, 654, 846, 997, 999, 16.5262, 100, 47.0578, 55.9464, 79.1087, 28.3014, 15.5285, 38.5213, 78.019, 100 \\ 
17096.2848 & 0.0868 & 3, 6, 154, 290, 402, 654, 846, 997, 999, 16.3784, 100, 47.0252, 55.9705, 79.0456, 28.1314, 15.4734, 38.434, 77.9669, 100 \\ 
23153.3221 & 0.0810 & 4, 148, 152, 344, 387, 589, 626, 929, 999, 100, 36.0277, 33.1264, 19.2331, 56.1436, 70.1146, 45.7855, 66.9093, 3.2387, 49.5776 \\ 
48330.0397 & 0.0800 & 48, 64, 107, 111, 312, 316, 616, 968, 999, 23.274, 21.9881, 82.6938, 31.9064, 51.8522, 46.7522, 7.3134, 7.885, 17.7037, 28.4997 \\ 
69227.3495 & 0.0704 & 81, 85, 140, 144, 375, 379, 662, 992, 999, 15.8308, 3.3201, 82.027, 31.5546, 49.6736, 44.1554, 5.5301, 5.1966, 1, 27.7097 \\ 
175425.0305 & 0.0640 & 4, 77, 81, 264, 363, 515, 519, 862, 999, 99.714, 15.4266, 12.8579, 1, 24.5412, 44.4314, 3.6139, 44.0644, 1, 78.2006 \\ 
214413.9704 & 0.0589 & 62, 81, 104, 108, 330, 333, 640, 984, 999, 6.7505, 5.1224, 84.4933, 28.4786, 49.5277, 47.2393, 8.3612, 1, 1, 25.3833 \\ 
215701.1479 & 0.0587 & 62, 81, 104, 108, 330, 333, 640, 984, 999, 6.518, 4.6684, 84.7306, 27.9346, 49.2162, 47.3169, 7.9741, 1, 1, 24.9831 \\ 
216016.0265 & 0.0586 & 62, 81, 104, 108, 330, 333, 640, 984, 999, 6.4985, 4.6521, 84.776, 27.845, 49.1622, 47.3666, 7.8545, 1, 1, 24.9409 \\ 
216205.5941 & 0.0586 & 62, 81, 104, 108, 330, 333, 640, 984, 999, 6.4677, 4.5925, 84.8079, 27.772, 49.1204, 47.3778, 7.8014, 1, 1, 24.8877 \\ 
216872.8269 & 0.0585 & 62, 81, 104, 108, 330, 333, 640, 984, 999, 6.4942, 4.7294, 84.8796, 27.6636, 49.0478, 47.5367, 7.5073, 1, 1, 24.9001 \\ 
219061.4873 & 0.0569 & 62, 85, 104, 108, 330, 333, 644, 988, 999, 5.9005, 3.4856, 85.3872, 26.444, 48.3602, 47.5678, 6.8553, 1, 1, 23.9115 \\ 
237536.5158 & 0.0568 & 3, 6, 83, 198, 443, 708, 861, 997, 999, 1, 100, 7.5769, 26.1086, 38.138, 1, 1, 19.4766, 67.3284, 100 \\ 
340407.3056 & 0.0512 & 70, 74, 77, 81, 359, 363, 666, 996, 999, 3.5341, 1, 91.1333, 24.8442, 43.6103, 37.2859, 1, 1, 1, 18.4452 \\ 
342099.4535 & 0.0477 & 73, 77, 81, 85, 341, 345, 651, 996, 999, 6.5932, 3.5406, 86.5368, 29.1274, 48.2486, 42.6629, 1, 1, 1, 24.0753 \\ 
378791.5280 & 0.0460 & 77, 81, 85, 89, 346, 350, 650, 996, 999, 1, 1, 80.9047, 32.3635, 44.5046, 40.2576, 1, 1, 1, 19.1952 \\ 
381352.5337 & 0.0422 & 73, 77, 81, 85, 333, 337, 651, 996, 999, 1, 1, 83.4641, 28.5675, 46.4756, 42.5143, 1, 1, 1, 17.77 \\ 
397730.6830 & 0.0355 & 43, 70, 74, 77, 300, 304, 650, 996, 999, 1, 1, 94.8855, 4.1151, 32.8482, 47.1522, 1, 1, 1, 4.5378 \\ 
398055.8357 & 0.0352 & 43, 70, 74, 77, 300, 304, 650, 996, 999, 1, 1, 95.6221, 2.6625, 31.7785, 46.9191, 1, 1, 1, 3.3486 \\ 
398151.7738 & 0.0351 & 43, 70, 74, 77, 300, 304, 650, 996, 999, 1, 1, 95.7505, 2.3605, 31.5771, 46.9199, 1, 1, 1, 3.0926 \\ 
398347.1561 & 0.0350 & 43, 70, 74, 77, 300, 304, 650, 996, 999, 1, 1, 95.9524, 1.886, 31.2606, 46.9211, 1, 1, 1, 2.6904 \\ 

\end{longtable}

\begin{figure}
    \centering
    \begin{minipage}[b]{0.75\textwidth}
        \includegraphics[width=\textwidth]{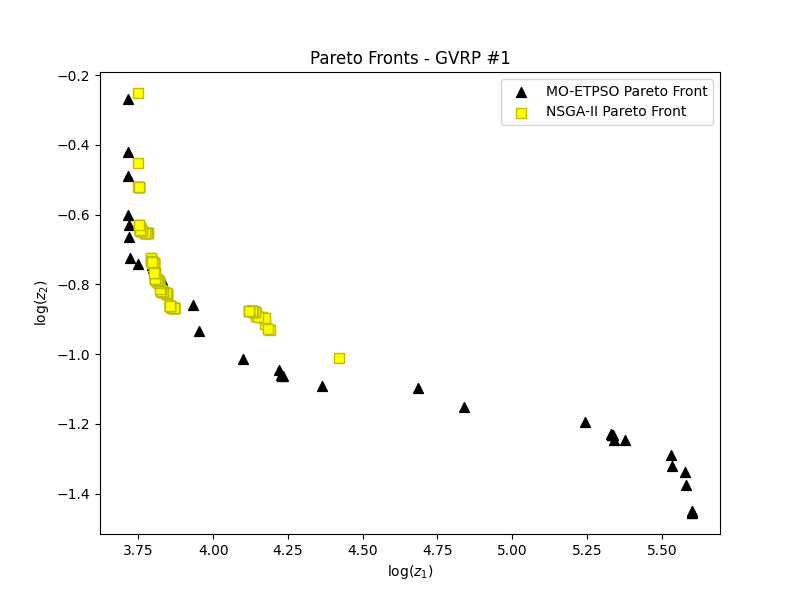}
        \caption{MO-ETPSO and NSGA-II Pareto Fronts for GVRP \#1}
        \label{fig8}
    \end{minipage}
    \hfill % optional for extra horizontal spacing
    \begin{minipage}[b]{0.75\textwidth}
        \includegraphics[width=\textwidth]{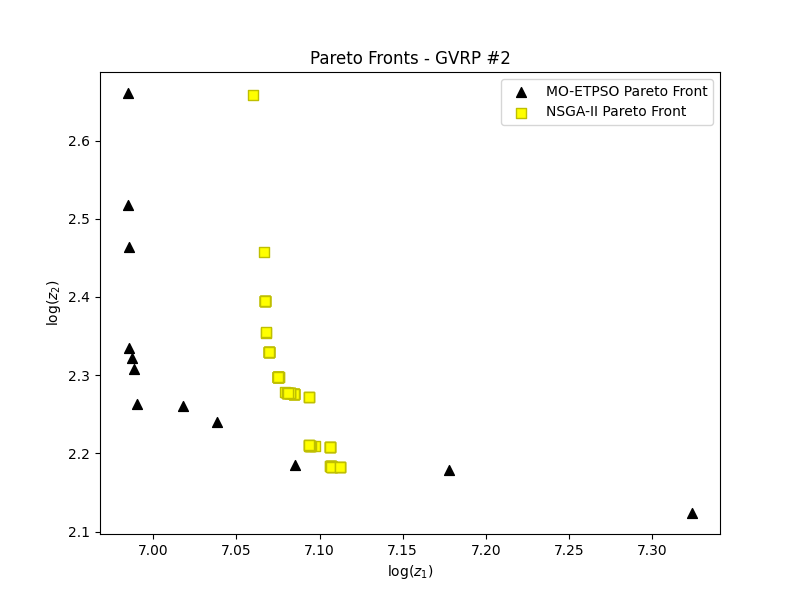}
        \caption{MO-ETPSO and NSGA-II Pareto Fronts for GVRP \#2}
        \label{fig9}
    \end{minipage}
\end{figure}

Figures \ref{fig8} and \ref{fig9} show the Pareto front of MO-ETPSO and NSGA-II at the end of the 100 iterations for Problems \#1 and \#2, respectively. The log scale was taken in order to observe the fronts better. For Problem \#1, it is possible to observe that the MO-ETPSO Pareto front has a great part of it below the NSGA-II. For Problem \#2, the MO-ETPSO Pareto front is substantially below the NSGA-II. This means MO-ETPSO outperforms NSGA-II, especially for highly dimensional problems.

Moreover, two parameters are also considered for a more quantified assessment of this comparison: the normalized hypervolume (HV) and the convergence score. The first parameter is defined in (\ref{eq16}) (considering the objectives space bi-dimensional):

\begin{equation}
    \bar{HV}_{\text{iter}} = \frac{1}{R}\sum_{i=1}^{R}HV_{\text{iter}}^{(i)}
\end{equation}
\begin{equation}
    HV_{\text{iter}}^{(i)} = \prod_{m=1}^2 1 - \bar{Z}_1^{(m)} + \sum_{i=2}^k \prod_{m=1}^2 1 - \bar{Z}_i^{(m)} - (1 - \bar{Z}_{i-1}^{(m)}) \cdot \max (0, m-1)
    \label{eq16}
\end{equation}

Where:

\begin{equation}
    \bar{Z}_i^{(m)} = \frac{\log(\textbf{MAX}^{(m)}) - \log(Z_i^{(m)})}{\log(\textbf{MAX}^{(m)}) - \log(\textbf{MIN}^{(m)})} 
    \label{eq17}
\end{equation}
\begin{equation}
    Z_i: Z_i^{(2)} \geq Z_{i+1}^{(2)}, \forall i \neq k
    \label{eq18}
\end{equation}

Where \textbf{MAX} and \textbf{MIN} are the maximum and minimum values registered in all simulations in the objectives space. 

\begin{figure}[ht]
    \centering
    \begin{minipage}[b]{0.75\textwidth}
        \includegraphics[width=\textwidth]{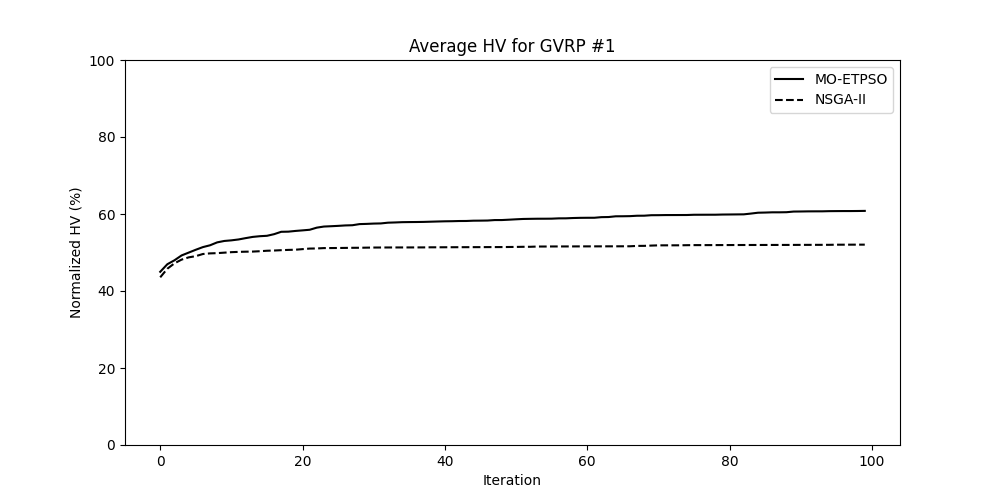}
        \caption{MO-ETPSO and NSGA-II Averaged HV for GVRP \#1}
        \label{fig10}
    \end{minipage}
    \hfill % optional for extra horizontal spacing
    \begin{minipage}[b]{0.75\textwidth}
        \includegraphics[width=\textwidth]{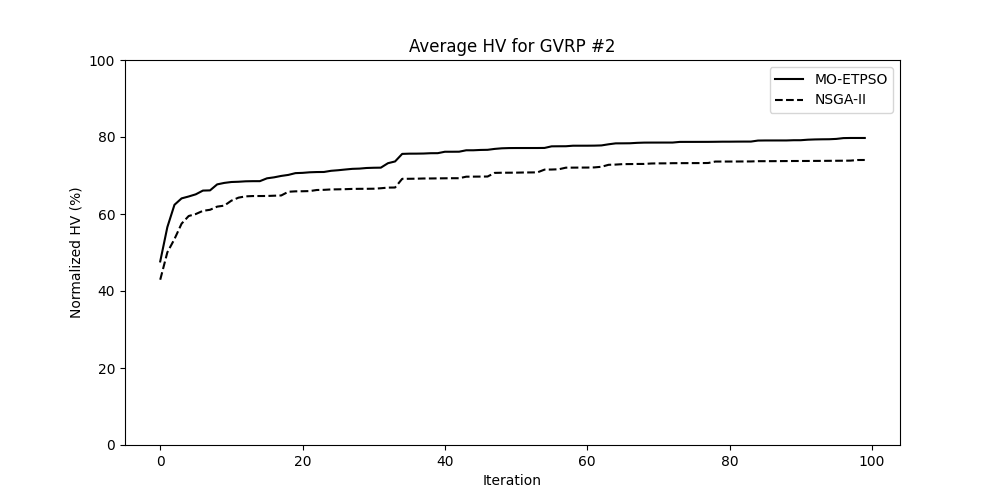}
        \caption{MO-ETPSO and NSGA-II Averaged HV for GVRP \#2}
        \label{fig11}
    \end{minipage}
\end{figure}

Figures \ref{fig10} and \ref{fig11} are the plots of the normalized HV along the iterations for both algorithms MO-ETPSO and NSGA-II. In Problem \#1, it is possible to observe that both algorithms start on the same percentage of HV, but MO-ETPSO can find more solutions by exploiting continuous search subspaces, where NSGA-II is trapped in the local optimum at the end of less than 20 iterations. In Problem \#2, MO-ETPSO completely outperforms NSGA-II due to its relatively high normalized HV curve compared to the same curve for NSGA-II.

The second parameter is intended to measure the capacity of the algorithm to find better solutions for the improvement of the Pareto front. The Convergence Score (CS) can go from 0 to 100\%: 100\% means that new solutions can still be found, while 0\% means that the algorithm is over because a new solution was not found in the last 100 iterations. This means that the higher the CS, the greater the probability of the algorithm finding new solutions.

It is calculated as follows:
\begin{equation}
    \bar{CS}_{\text{iter}} = \frac{1}{R}\sum_{i=1}^{R}CS_{\text{iter}}^{(i)}
\end{equation}
\begin{equation}
    CS_{\text{iter}}^{(i)} = 
    \begin{cases} 
    1 & \text{if } EV_{\text{iter}}^{(i)} > EV_{\text{iter}-1}^{(i)} \text{ or } \text{iter} = 1 \\
    CS_{\text{iter}-1}^{(i)} - 0.01 & \text{otherwise}
    \end{cases}
\end{equation}

\begin{figure}[ht]
    \centering
    \begin{minipage}[b]{0.75\textwidth}
        \includegraphics[width=\textwidth]{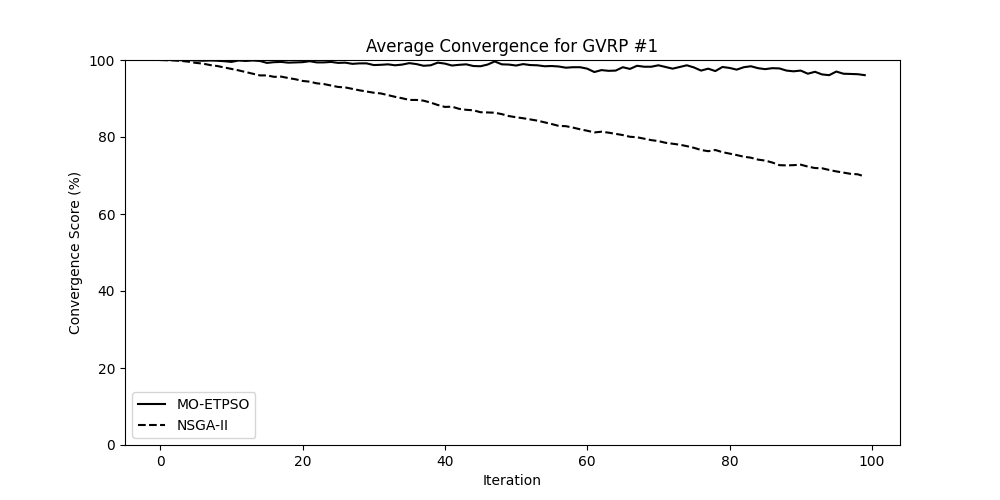}
        \caption{MO-ETPSO and NSGA-II Averaged Convergence Score for GVRP \#1}
        \label{fig12}
    \end{minipage}
    \hfill % optional for extra horizontal spacing
    \begin{minipage}[b]{0.75\textwidth}
        \includegraphics[width=\textwidth]{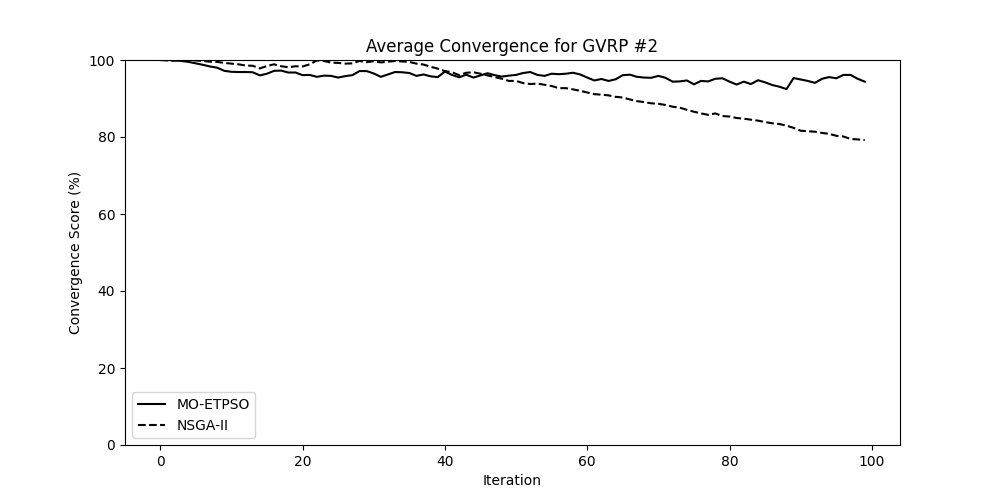}
        \caption{MO-ETPSO and NSGA-II Averaged Convergence Score for GVRP \#2}
        \label{fig13}
    \end{minipage}
\end{figure}

Figures \ref{fig12} and \ref{fig13} show the average convergence score for both MO-ETPSO and NSGA-II, for the problems \#1 and \#2, respectively. It is possible to observe that MO-ETPSO can still find more solutions at the end of 100 iterations, whereas NSGA-II is not able to do so at the end of about 40 iterations. This is especially true for the first problem, where the CS is still higher than 90\% with MO-ETPSO and below 70\% for NSGA-II.

The MO-ETPSO results shown here may have some potential explanations. PSO itself has two useful advantages compared to GA. One of them is that the exploitation of sub-spaces is performed more efficiently, despite the fact that sometimes it leads to local optima without the possibility of further exploration. This problem is solved by using the randomized techniques of simply one particle at a time in order to improve exploration, leveraging then the advantage of effective exploitation of PSO. The second advantage is related to the existence of an individual level of reference, the 'pbest,' that allows particles to not blindly follow the same objective as the other particles, maintaining a level of diversity in the individual objectives. This does not happen in GA. MO-ETPSO effectively picks on this advantage of PSO but still leverages the crowding distance and rank techniques of NSGA-II.

\subsection{Computational Efficiency}
All the experiments were conducted on a standardized computational platform to ensure consistency and reproducibility of the results. The hardware configuration consisted of 12th Gen. Intel(R) Core(TM) i9-12900KF 3.20 GHz, with 128GB of RAM. Each algorithm was implemented in Python 3.12.

The computational efficiency was evaluated based on the average of 5 runs for each algorithm and problem. For Problem \#1, MO-ETPSO took $2.88 \pm 0.03$ seconds, and NSGA-II took $3.00 \pm 0.04$ seconds. For Problem \#2, MO-ETPSO took $72.49 \pm 0.04$, and NSGA-II took $143.42 \pm 0.25$.

\subsection{Limitations}

This study has some limitations. First, a limited set of experiments has been done. Neither the effect of the hyperparameters nor other problems were tested. According to the No Free Lunch Theorem \cite{R039}, there might be problems where MO-ETPSO does not perform as well as NSGA-II, and the author acknowledges those sets of problems are still to be understood and how to change the algorithm so MO-ETPSO can have the desired adaptability. Future research is still to be done in this regard.

Moreover, MO-ETPSO could be adapted so that multiple different objectives can be attributed to the particles, promoting more swarm-related efficiency while sharing information between particles at the same time.

\section{Conclusions and Future Works}

This study introduces a simplified Elitist Particle Swarm Optimization (MO-ETPSO) algorithm, specially designed to deal with multi-objective optimization problems. By integrating swarm intelligence principles, swarm memory, and elitism, MO-ETPSO demonstrates a novel approach to approach complex solution spaces. The algorithm performance is evaluated through the GVRP, which showcases significant improvements in convergence rates and the quality of the solution set when compared to the known NSGA-II algorithm. These results highlight MO-ETPSO's potential to offer a more efficient and effective framework for solving real-world optimization problems.

 MO-ETPSO Pareto front was compared to the NSGA-II Pareto front, and it was evident to have an improvement, not only in terms of diversity but also most of the first Pareto front are dominant compared to those from the second one. By introducing random sampling techniques along the iterations, and by leveraging the concept of 'pbest,' the algorithm is capable of effectively balancing exploration and exploitation within the search space. Additionally, the incorporation of a memory mechanism ensures the retention of high-quality solutions across iterations, further contributing to its efficacy.

The algorithm's performance is notably influenced, however, by the initial configuration of its parameters, which can pose challenges in terms of adaptability and generalization across different types of optimization problems. This sensitivity necessitates further research to develop adaptive parameter tuning methods, ensuring this simplified MO-ETPSO's robustness and versatility in a broader array of application scenarios.

The author suggests future works to address the hyperparameter sensitivity analysis and that MO-ETPSO be adapted in multiple variants according to the different problems being solved. These directions promise to extend the applicability of MO-ETPSO, making it a more powerful and accessible tool for researchers in the field of multi-objective optimization. The author also hopes MO-ETPSO can be applied to real-world problems in the future.

\bibliography{references}

\end{document}